\documentclass{article}

\usepackage[a4paper, margin=1in]{geometry}
\usepackage[numbers]{natbib} 
\usepackage[utf8]{inputenc}
\usepackage[LGR, T1]{fontenc}
\usepackage{lmodern}
\usepackage{fancyvrb}
\usepackage{listings}
\usepackage{float}
\usepackage{comment}
\usepackage{authblk}
\usepackage[hyphens,spaces,obeyspaces]{url}
\usepackage{array}
\usepackage{makecell}
\usepackage{csquotes}
\usepackage[dvipsnames]{xcolor}
\usepackage{parskip}
\usepackage{etoolbox}
\usepackage{paralist} 
\usepackage{enumitem}
\usepackage{microtype}
\usepackage{graphicx}
\usepackage{booktabs}
\usepackage{amsmath}
\usepackage{amssymb}
\usepackage{mathtools}
\usepackage{amsthm}
\usepackage{mathrsfs} 
\usepackage{todonotes}
\usepackage[breaklinks,colorlinks]{hyperref}
\usepackage[capitalize,noabbrev]{cleveref}
\usepackage{bm}

\newcommand{\seq}[1]{\bm{#1}}
\newcommand{\pr}[1]{#1^\ast}
\DeclareMathOperator*{\argmax}{arg\,max}

\newcommand{\gpt}{\textsc{GPT}}
\newcommand{\gptfour}{\textsc{GPT-4}}
\newcommand{\chatgpt}{\textsc{ChatGPT}}
\newcommand{\gptthree}{\textsc{GPT-3}}
\newcommand{\gpttwo}
{\textsc{GPT-2}}
\newcommand{\gptthreefive}{\textsc{GPT-3.5-turbo}}
\newcommand{\llama}{\textsc{LLaMA}}
\newcommand{\llamatwo}{\textsc{LLaMA2}}

\DeclareRobustCommand{\textvtt}[1]{%
  \begingroup
  \ttfamily
  \hyphenchar\font=`\- 
  \setlength{\spaceskip}{0.8em plus 0.4em minus 0.2em}%
  \setlength{\xspaceskip}{1em plus 0.4em minus 0.2em}%
  #1%
  \endgroup
}

\makeatletter
\DeclareRobustCommand\vttfamily{%
  \not@math@alphabet\vttfamily\relax
  \fontfamily{cmvtt} 
  \selectfont
}
\DeclareTextFontCommand{\textvtt}{\vttfamily}
\makeatother

\setlist[itemize]{align=parleft,left=0pt..1em}

\definecolor{darkcolor}{RGB}{127,0,85}
\colorlet{numb}{magenta!60!black}
\lstdefinelanguage{json}{
    basicstyle=\fontsize{8.6}{11}\vttfamily,
    commentstyle=\color{black},
    stringstyle=\color{darkcolor},
    showstringspaces=false,
    breaklines=true,
    breakatwhitespace=true,
    frame=lines,
    string=[s]{"}{"},
    comment=[l]{:\ "},
    morecomment=[l]{:"},
    literate=
        *{0}{{{\color{numb}0}}}{1}
         {1}{{{\color{numb}1}}}{1}
         {2}{{{\color{numb}2}}}{1}
         {3}{{{\color{numb}3}}}{1}
         {4}{{{\color{numb}4}}}{1}
         {5}{{{\color{numb}5}}}{1}
         {6}{{{\color{numb}6}}}{1}
         {7}{{{\color{numb}7}}}{1}
         {8}{{{\color{numb}8}}}{1}
         {9}{{{\color{numb}9}}}{1}
}

\hypersetup{
    urlcolor=TealBlue!40!black,
    citecolor=TealBlue!40!black,
    linkcolor=TealBlue!40!black,
    anchorcolor=TealBlue!40!black,
    citecolor=TealBlue!40!black,
    filecolor=TealBlue!40!black,
    menucolor=TealBlue!40!black,
    runcolor=TealBlue!40!black,
}

\ifundef{\abstract}{}{\patchcmd{\abstract}%
    {\quotation}{\quotation\noindent\ignorespaces}{}{}}

\title{Large Language Models for Mathematicians}
\author{Simon Frieder\thanks{Department of Computer Science, University of Oxford, Oxford, UK}, Julius Berner\thanks{Department of Computing and Mathematical Sciences, Caltech, Pasadena, US}, Philipp Petersen\thanks{Faculty of Mathematics and Research Network Data Science, University of Vienna, Vienna, Austria}, Thomas Lukasiewicz\thanks{Institute of Logic and Computation, Vienna University of Technology, Vienna, Austria, and  Department of Computer Science, University of Oxford, Oxford, UK}}
\date{}

\begin{document}

\maketitle
\begin{abstract}
Large language models (LLMs) such as \chatgpt~have received immense interest for their general-purpose language understanding and, in particular, their ability to generate high-quality text or computer code. For many professions, LLMs represent an invaluable tool that can speed up and improve the quality of work. In this note, we discuss to what extent they can aid professional mathematicians. We first provide a mathematical description of the transformer model used in all modern language models. Based on recent studies, we then outline best practices and potential issues and report on the mathematical abilities of language models. Finally, we shed light on the potential of LLMs to change how mathematicians work.
\end{abstract}

\section{Introduction}

\emph{Large language models} (LLMs) recently revolutionized the field of natural language processing, bringing forth unparalleled progress for applications such as conversational bots or text generation. Advances in deep-learning algorithms, computational resources, as well as substantial engineering efforts, allowed to train language models with billions of parameters and on massive datasets, such as the $800$GB large \emph{The Pile}~\cite{gao2020pile}. Being trained on increasingly larger datasets, language models infer essential aspects of the structure of language\footnote{We note that language models work best in English since the most extensive datasets, including those that contain mathematics, are available in English.}, paving the way from task-specific to general-purpose language understanding. As such, these models can also parse mathematical questions and solve (sometimes surprisingly complex) mathematical problems and tasks---however, they also make embarrassing mistakes.

Let us have a look at two examples that \chatgpt~(web version of 30-January-2023) produced\footnote{The first example can by now be correctly answered by the \gptfour~model~\cite{openai2023gpt}, which was since released. We explain the acronym \enquote{\gpt} in Section~\ref{sec: overview}.}. 
Both examples below were taken from \cite{frieder2023mathematical}; the input to the model, as well as its output, consists of \LaTeX~code, which we have compiled below for better readability:

\begin{lstlisting}[language=json, mathescape=true, escapeinside={(*}{*)}]
"prompt": "Find a function $f : \mathbb{R} \to \mathbb{R}$ that is continuous at precisely one point.",
"output": "One example of such a function is (*$$f(x) = \begin{cases}x^2 &\text{if } x \neq 0 \\ 1 &\text{if } x = 0. \end{cases}$$*)This function is continuous at $x=0$ but not at any other point."
\end{lstlisting}

\begin{lstlisting}[language=json, mathescape=true, escapeinside={(*}{*)}]  
"prompt": "Let $\mu$ be a measure, $\|f\|_\infty = \inf\{M : \mu(\{x : |f(x)| > M\}) = 0\}$ and  $\|g\|_1 = \int |g| d\mu$. Prove that $\int |f g |d\mu \leq \|g\|_1 \|f\|_\infty$.",
"output": "Let $M = \|f\|_\infty$ and $\epsilon > 0$. Then, by the definition of $\|f\|_\infty$, we have $\mu(\{x : |f(x)| > M + \epsilon\}) = 0$. Therefore, (*$$\int |f g| d\mu = \int_{|f| \leq M + \epsilon} |f g| d\mu + \int_{|f| > M + \epsilon} |f g| d\mu \leq (M + \epsilon) \int |g| d\mu + \int_{|f| > M + \epsilon} |f| |g| d\mu.$$*)The second term on the right side is zero because $\mu({x : |f(x)| > M + \epsilon}) = 0$. So we have $\int |f g| d\mu \leq (M + \epsilon) \int |g| d\mu = (M + \epsilon) \|g\|_1$. As $\epsilon$ can be arbitrarily small, we can let it go to zero and we get $\int |f g| d\mu \leq M \|g\|_1 = \|f\|_\infty \|g\|_1$. Hence proved."
\end{lstlisting}
The first response is nonsensical on all levels, whereas the second is correct and helpful. With these examples in mind, we might ask ourselves:

\begin{center}
\emph{How can large language models assist human mathematicians in their work?}
\end{center}

To address this question, this article proceeds in the following way: First, we provide an overview of modern language models. We clarify the theory of how language models work, how their main building block---the \emph{transformer} architecture---is set up, and why these models can perform sufficient mathematics to assist mathematicians in their daily work. Understanding this architecture will also highlight how an LLM produces an answer to a mathematical question -- which differs tremendously from how a mathematician arrives at an answer. 
Then, we present empirical evidence attesting to the abilities of language models, in particular state-of-the-art models, such as \chatgpt~and \gptfour. 
We end with an outlook on the potential future impacts on mathematicians and mathematics in general.

\section{Overview of Modern Language Models}
\label{sec: overview}

The concept of language models has a long history. One of the pioneering achievements, dating back to the year 2000, presented one of the initial instances of what we now refer to as \emph{word embeddings} within the framework of neural networks~\cite{bengio2000neural}; see Section~\ref{sec: technical background} for a definition.

Most previous approaches were rooted in estimating probabilities over \emph{trigrams} (or, more general, $n$-grams). An $n$-gram is a sequence of $n$ adjacent elements from a string of word pieces, so-called \emph{tokens}, which could be syllables, letters, words, or base pairs according to the context. In the sentence \enquote{\texttt{The quick brown fox jumps over the lazy dog}}, the sequence \enquote{\texttt{quick brown fox}} is an example of a trigram. Models based on $n$-grams had severe limitations: For example, if a trigram does not appear in the training corpus (or contains words that were not in the vocabulary of the corpus), no meaningful way of estimating its probability existed. By using a form of word embeddings, these problems are circumvented. The model proposed by~\cite{bengio2000neural} dominated all other pure $n$-gram models. The authors note that improvements can be made regarding the \enquote{\emph{architecture,
computational efficiency, and taking advantage of prior knowledge}}. 

The introduction of the \emph{transformer architecture}~\cite{vaswani2017attention} in 2017 marked the most striking advancement in terms of neural network architectures: On the one hand, the attention mechanism modeled the structure of the language more faithfully; on the other hand, it was an architecture that was easily parallelizable on modern hardware (see Section~\ref{sec: technical background} for details). This led to a series of further milestones and improvements: In 2018, the \emph{Bidirectional Encoder Representations from Transformers} (\textsc{BERT}) model~\cite{devlin2018bert} was introduced, a successor to the original transformer, which inspired a vast number of successors on its own, such as \textsc{RoBERTa}~\cite{liu2019roberta}, or \textsc{DistilBERT}~\cite{sanh2019distilbert}. \textsc{BERT} (and its successors) were notable because classical pipelines 
(e.g., defining text representations, carrying out parts-of-speech tagging) were all subsumed by \textsc{BERT}-type models~\cite{tenney2019bert}, which could easily be fine-tuned to specific tasks. At roughly the same time as the \textsc{BERT} model, the \emph{Generative Pre-Trained Transformer} (GPT) model was introduced by OpenAI~\cite{radford2018improving}. This was a further variation on the original transformer architecture and is the first version of the model that underlies \chatgpt, which was released in 2022~\cite{openai2022chatgpt}, and is closely related to \textsc{InstructGPT}~\cite{ouyang2022training}.

\begin{figure}[htb]
    \centering
    \includegraphics[width=\linewidth]{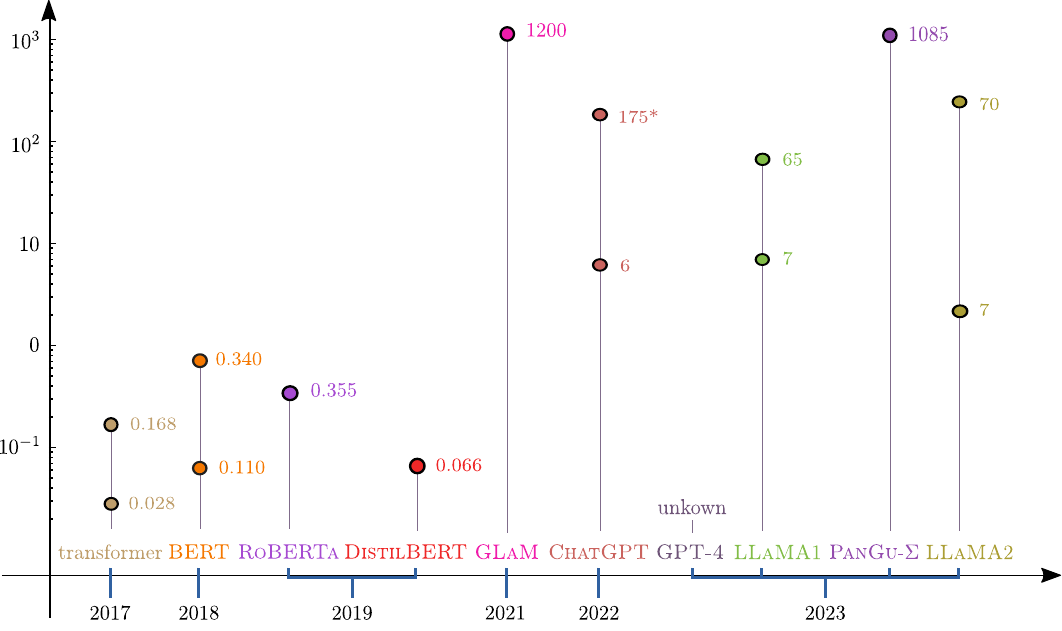}
    \caption{A selection of representative, modern language models is presented along with their parameter counts (in billions), which are displayed above. The y-axis is a log-axis, and model ranges are displayed (two horizontal dots), where available, for each model. We observe that a wide range of parameters appears, between 28 million and 1.2 trillion.  For \chatgpt, exact parameter counts are not available but are taken from \textsc{InstructGPT}, which is a sibling model on which \chatgpt{} is based. For \gptfour, parameter counts are not available.}
    \label{fig: timeline}
\end{figure}

The last milestone consists of the \llama~\cite{touvron2023llama1} and \llamatwo~models~\cite{touvron2023llama} introduced in 2023, months after \gptfour~\cite{openai2023gpt}. Their importance lies in being the first publicly released models, the code and weights of which were easily accessible and rivaled the performance of \gptfour; in the technical report associated with \gptfour{} it is stated: \enquote{\emph{this report
contains no further details about the architecture (including model size), hardware, training compute,
dataset construction, training method, or similar}}. The \llama{} models led to a democratization of language models and to a large number of further successors, such as Stanford's \textsc{Alpaca}\footnote{\url{https://github.com/tatsu-lab/stanford_alpaca}} model, or the \textsc{Vicuna}\footnote{\url{https://lmsys.org/blog/2023-03-30-vicuna/}} model, which have since been used in a wide array of contexts. As these models evolved, the number of their parameters, as well as the sizes of the dataset on which they were trained, kept increasing, from the order of millions of parameters (in case of \cite{bengio2000neural, vaswani2017attention, devlin2018bert}), to billions~\cite{touvron2023llama, touvron2023llama1}, to trillions~\cite{du2022glam, ren2023pangu}, see Figure~\ref{fig: timeline}. While the main trend indicates increasing model sizes, there is a countertrend to make the models smaller while retaining the performance. The \textsc{DistilBERT} model is an example of this. Scaling the size of the architectures and the amount of training data enabled unprecedented capabilities for the resulting LLMs, eliminating the need for fine-tuning for specific tasks.

\section{Technical Background}
\label{sec: technical background}

\begin{figure}
    \centering
    \includegraphics[width=\linewidth]{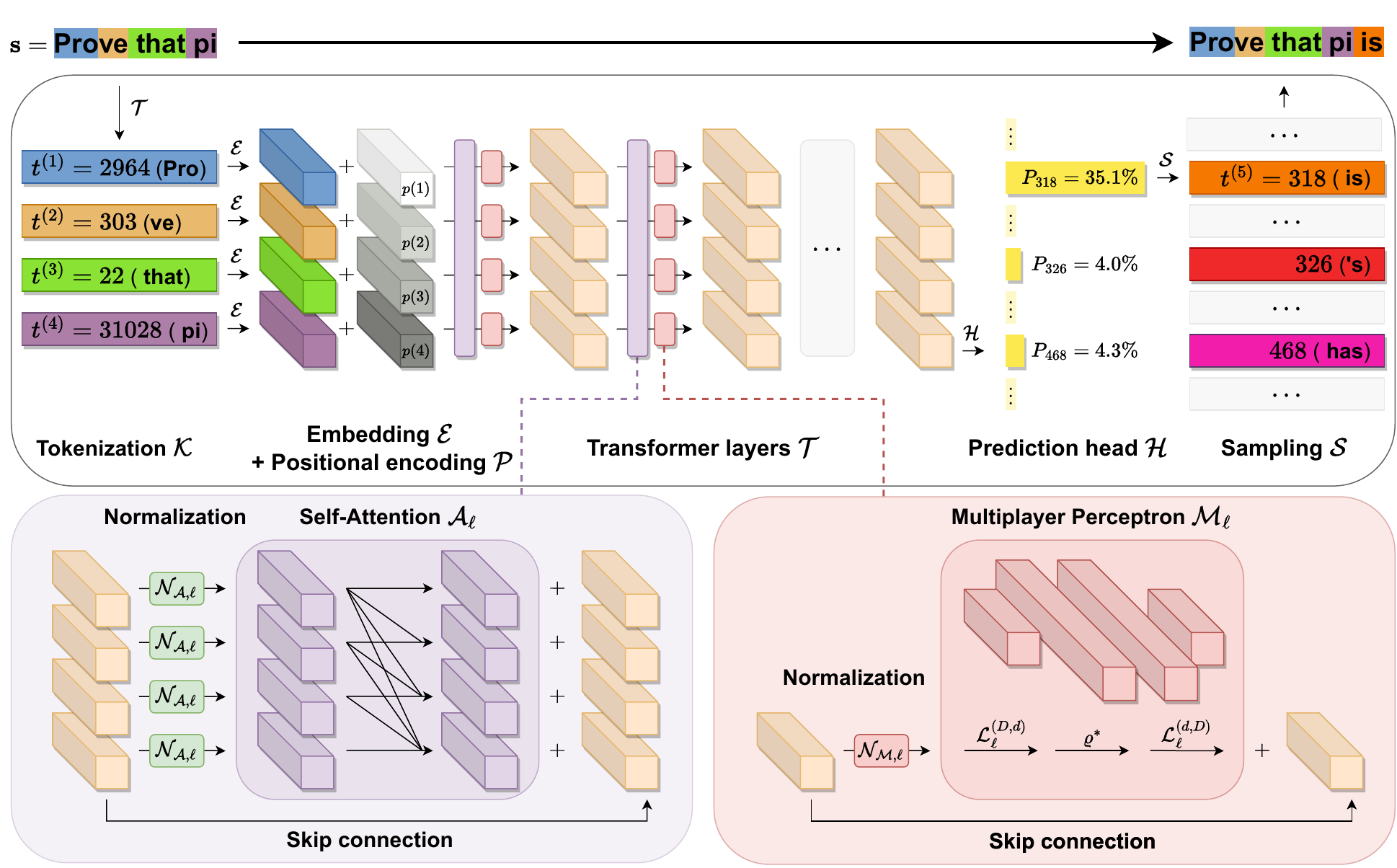}
    \caption{Illustration of the operations of an LLM for the input text \enquote{\texttt{Prove that pi}}. The token indices, as well as the probabilities for the next token, are taken from \gpttwo~\cite{radford2019language} using the implementation in the \texttt{transformers} library~\cite{wolf-etal-2020-transformers}. The highest probability for the next token is assigned to \texttt{318} corresponding to the word \enquote{\texttt{is}}.}
    \label{fig:transformer}
\end{figure}

In the following, we seek to give a brief introduction to the inner workings of LLMs. We refer to \cite{zhao2023survey,min2023recent} for surveys and further details. We do not strive to present state-of-the-art models and techniques in natural language processing but focus on a conceptual understanding of the functionality of LLMs. In particular, we will restrict the presentation to one of the most popular architectures, the transformer~\cite{vaswani2017attention} (see Section~\ref{sec: overview} for context regarding the importance of this model). Our description below is a simplified, mathematical summary loosely based on the (open-source) code\footnote{\url{https://github.com/openai/gpt-2}} of \gpttwo~\cite{radford2019language}, see also Figure~\ref{fig:transformer} for an overview.

\subsection{Transformer Architecture}

Let us explore how a transformer architecture predicts text based on some provided input, often referred to as \emph{prompt}. In line with successful models, such as the \gpt~and \llama~series, we focus on the setting, where the model iteratively predicts the next word pieces (i.e., tokens) based on a given sequence of tokens. 
This procedure is coined \emph{autoregressive} since the prediction of new tokens is only
based on previous tokens.
Such conditional sequence generation tasks using autoregressive transformers are often referred to as \emph{decoder-only} settings. 
Although there is a maximum \emph{context length} in practice, we will work with sequences of arbitrary length for ease of presentation. We define the shorthand notation $\pr{S}\coloneqq \bigcup_{n\in\mathbb{N}} S^n$ for a set $S$ to denote the set of sequences $\seq{s}=(s^{(i)})_{i=1}^n\subset S$ with arbitrary length $n\in\mathbb{N}$. For a function $\mathcal{F}\colon S_1 \to S_2$, we denote by $\pr{\mathcal{F}}\colon \pr{S_1} \to \pr{S_2}$ the \emph{entrywise} applied mapping given by
\begin{equation}
    \pr{\mathcal{F}}(\seq{s}) \coloneqq (\mathcal{F}(s^{(i)}))_{i=1}^n.
\end{equation} 

\paragraph{Tokenization $\mathcal{K}$.}

First, we want to clarify how we define word pieces, i.e., tokens. Mathematically speaking, we seek an injective mapping $\mathcal{K}\colon\pr{A} \to \pr{T}$ from the given text, i.e., a sequence $\seq{a}=(a^{(i)})_{i=1}^N$ of characters in an alphabet $A$ to a sequence of $n\le N$ tokens $(t^{(i)})_{i=1}^n$, where typically $T\coloneqq \{1,2,\dots,M\}$.

To represent text on a computer, we could encode every character $a^{(i)}$ individually, similar to \emph{Unicode}. While this would lead to a small vocabulary $T$ of tokens, it yields long sequences where individual tokens do not capture any linguistic information. For LLMs, one often employs \emph{subword tokenization}~\cite{sennrich2015neural}, which creates a vocabulary of subwords by analyzing large text corpora and iteratively merging frequently occurring sequences of characters. Akin to a compression problem, one balances the length $n$ of the sequences and the size\footnote{The \llama~and \llamatwo~models employ vocabularies $T$ with $M=32000$ tokens~\cite{touvron2023llama}. \gpttwo~uses $M=50257$, and other models in the \gpt~series, e.g., \gptthreefive~and \gptfour, even use $M=100277$ tokens, see~\url{https://github.com/openai/tiktoken}.} $M$ of the vocabulary. As an example, the \gptfour~tokenizer\footnote{See \url{https://platform.openai.com/tokenizer}.} splits the word \enquote{discontinuity} into the subwords
\enquote{dis} (prefix),
\enquote{contin} (subword capturing the root of \enquote{continuous}), and
\enquote{uity} (suffix). Another example can be found in Figure~\ref{fig:transformer}.

\paragraph{Embedding $\mathcal{E}$.} To use these tokens (given by the indices of the subwords in the vocabulary) in a neural network, we embed each token $t^{(i)}$ into the same Euclidean space $E\coloneqq \mathbb{R}^d$. Intuitively, we seek a map $\mathcal{E}\colon T \to E$, such that the distance $\| \mathcal{E}(t^{(i)}) - \mathcal{E}(t^{(j)})\|$ corresponds to the linguistic similarity of the subwords represented by the tokens $t^{(i)}$ and $t^{(j)}$. In practice, such an embedding is often initialized with a sequence of $M$ random initial embeddings and learned jointly with the transformer model from data. 

\paragraph{Positional Encoding $\mathcal{P}$.}
Since $\mathcal{E}$ operates on each token $t^{(i)}$ independently, the embeddings $\mathcal{E}(t^{(i)})$ do not contain information on the position $i$ of the (sub)words within a sentence\footnote{In some settings, where the transformer architecture is permutation invariant, positional encodings are strictly necessary. In our decoder-only setting, this is not the case~\cite{haviv2022transformer}; however, the encodings still seem to improve performance}. 
Thus, one typically adds so-called \emph{positional encodings}, which can be described by a mapping $\mathcal{P}\colon \pr{E} \to \pr{E}$. A commonly used choice is of the form 
\begin{equation}
    \mathcal{P}((e^{(i)})_{i=1}^n) \coloneqq  (e^{(i)} + p(i))_{i=1}^n,
\end{equation}
where $p\colon\mathbb{N} \to E$ can be a prescribed injective function, e.g., a sinusoid~\cite{vaswani2017attention}, or learned (similar to the embedding $\mathcal{E}$)~\cite{radford2018improving}.

In summary, tokenization $\mathcal{K}$, followed by an application of $\mathcal{E}$ to each token and the positional encoding $\mathcal{P}$, maps the text $\seq{a}\in \pr{A}$ to a sequence of embeddings 
\begin{equation}
    \seq{e} \coloneqq \left(\mathcal{P} \circ \pr{\mathcal{E}} \circ \mathcal{K} \right)(\seq{a}) \in \pr{E}.
\end{equation}
where the length of $\seq{e}$ depends on $\seq{a}$ and the tokenization algorithm.
\paragraph{Transformer $\mathcal{T}$.} The transformer can be represented as a neural network $\mathcal{T}\colon \pr{E} \to \pr{E}$. It is trained to map a sequence of embeddings $\seq{e}$ to another sequence of the same length containing contextual information. Based on the desired autoregressive structure, where the prediction of the next token only depends on the previous tokens, we want the $i$-th element of $\mathcal{T}(\seq{e})$ to contain information about all the embeddings $(e^{(j)})_{j\le i}$, however, to be independent of $(e^{(j)})_{j>i}$.

The transformer is typically defined by a composition of $L\in\mathbb{N}$ blocks, consisting of \emph{self-attention} maps $\mathcal{A}_\ell$, entrywise applied \emph{normalizing layers} $\mathcal{N}_{\mathcal{A},\ell}$, $\mathcal{N}_{\mathcal{M},\ell}$, and \emph{feed-forward multiplayer perceptrons} $\mathcal{M}_\ell$, i.e.,
\begin{equation}
\label{eq:transformer}
    \mathcal{T} \coloneqq \left(\left(\mathrm{Id} + \pr{\mathcal{M}_L}\circ\pr{\mathcal{N}_{\mathcal{M},L}}\right) \circ \left(\mathrm{Id} + 
 \mathcal{A}_L\circ\pr{\mathcal{N}_{\mathcal{A},L}} \right)\right) \circ \dots \circ \left(\left(\mathrm{Id} + \pr{\mathcal{M}_1}\circ\pr{\mathcal{N}_{\mathcal{M},1}} \right) \circ \left( \mathrm{Id} + \mathcal{A}_1\circ\pr{\mathcal{N}_{\mathcal{A},1}}\right)\right).
\end{equation}
In the above, $\mathrm{Id}$ denotes the identity mapping, commonly known as a \emph{skip} or \emph{residual connection}, and the addition is understood entrywise. The indices of the layers $\mathcal{N}$, $\mathcal{M}$, and $\mathcal{A}$ in~\eqref{eq:transformer} indicate the use of different trainable parameters in each of the layers. Let us describe these layers in more detail below.

\paragraph{Layers: Normalization $\mathcal{N}$.}
The normalizing layer can be interpreted as a re-parametrization with a learnable mean and standard deviation to stabilize training. For instance, using \emph{layer normalization}
$\mathcal{N}\colon E\to E$, we compute
\begin{equation}
    \mathcal{N}(e) = \frac{\operatorname{diag}(s)}{\sigma} (e - \mu) + m,
\end{equation}
where $\mu = \frac{1}{d} \sum_{i=1}^d e_i$ and $\sigma^2=\frac{1}{d} \sum_{i=1}^d (e_i-\mu)^2$ are the mean and variance of $e\in E$, and $s,m\in E$
are learnable parameters~\cite{ba2016layer}.

\paragraph{Layers: Multilayer Perceptrons $\mathcal{N}$.}
The Multilayer perception (MLP) is a standard feed-forward neural network consisting of compositions of affine mappings and nonlinear activation functions. Let us define by $\mathcal{L}^{(m,n)}\colon \mathbb{R}^m \to \mathbb{R}^n$ an affine mapping $\mathcal{L}^{(m,n)} (x) \coloneqq Wx +b$, where the \emph{weight matrix} $W\in\mathbb{R}^{n \times m}$ and the \emph{bias vector} $b\in\mathbb{R}^{m}$ are learnable. Moreover, let $\varrho\colon\mathbb{R}\to\mathbb{R}$ be an activation function, e.g., the \emph{GELU activation function} $\varrho(x)\coloneqq x\, \Phi(x)$, where $\Phi$ is the standard Gaussian cumulative distribution function~\cite{hendrycks2016gaussian}. A typical MLP $\mathcal{M}\colon E\to E$ used in transformers is then given by 
\begin{equation}
\mathcal{M} \coloneqq \mathcal{L}^{(d,D)} \circ \pr{\varrho} \circ \mathcal{L}^{(D,d)},
\end{equation}
where $D\in\mathbb{N}$ with $D \ge d$.

\paragraph{Layers: Self-Attention $\mathcal{A}$.}
As can be seen in~\eqref{eq:transformer} and Figure~\ref{fig:transformer}, the self-attention layer $\mathcal{A}\colon \pr{E} \to \pr{E}$ is the only layer that combines embeddings of different tokens; in other words, it \emph{attends} to other tokens. 
Let us denote the input to the layer by $(e^{(i)})_{i=1}^n$ and focus on the $i$-th output. We first compute the (normalized) inner products 
\begin{equation}
s^{(i)}_{j} = \frac{1}{\sqrt{k}}\left\langle \mathcal{L}_{\mathrm{query}}^{(k,d)} (e^{(i)}), \mathcal{L}_{\mathrm{key}}^{(k,d)} (e^{(j)}) \right\rangle , \quad j =1, \dots,i,
\end{equation}
with given $k\in\mathbb{N}$. On a high level, we can interpret $\seq{s}^{(i)}=(s^{(i)}_{j})_{j=1}^i \subset \mathbb{R}$ as similarities between the embedding $\mathcal{L}_{\mathrm{query}}^{(k,d)} (e^{(i)})$ of the $i$-th token (i.e., the so-called \emph{query}) and the embeddings $\mathcal{L}_{\mathrm{key}}^{(k,d)}(e^{(j)})$ of the other tokens (i.e., \emph{keys}); to satisfy the autoregressive structure, we only consider $j \le i$.
To normalize $\seq{s}^{(i)}$ to probabilities, we can further use a \emph{softmax layer} $\operatorname{softmax}\colon \pr{\mathbb{R}} \to \pr{\mathbb{R}}$ 
given by
\begin{equation}
    \operatorname{softmax}(\seq{s}^{(i)})_j \coloneqq \frac{\exp\left({s^{(i)}_j}\right)}{\sum_{k=1}^i \exp\left( s_k^{(i)} \right)}, \quad j=1,\dots, i.
\end{equation}
We can now interpret $\operatorname{softmax}(\seq{s}^{(i)})_j$ as the probability for the $i$-th query to \enquote{attend} to the $j$-th key.
The self-attention layer $\mathcal{A}$ can then be defined as
\begin{equation}
   \mathcal{A}(\seq{e})_i  \coloneqq  \mathcal{L}^{(k,d)}\left( \sum_{j=1}^i \operatorname{softmax}(\seq{s}^{(i)})_j \mathcal{L}_{\mathrm{value}}^{(k,d)}(e^{(j)}) \right), \quad i=1,\dots,n,
\end{equation}
where the outputs of $\mathcal{L}_{\mathrm{value}}^{(k,d)}$ are often referred to as the \emph{values} of the token embeddings $e^{(j)}$, and where the learnable affine layer $\mathcal{L}^{(k,d)}$ maps the weighted average of values back to $E=\mathbb{R}^d$.

Note that in practice, one typically considers a sum of $h\in\mathbb{N}$ such attention layers (so-called \emph{heads}), each with dimension $k=d/h$~\cite{vaswani2017attention,liu2021multi}. Moreover, instead of 
considering vectors of variable length $i$, a mask enforces the autoregressive structure so that all operations can be efficiently batched.

\paragraph{Prediction Head $\mathcal{H}$.}
The \emph{prediction head} or \emph{un-embedding layer} can be represented as a mapping $\mathcal{H}\colon \pr{E} \to \Delta^{M}$, where 
\begin{equation}
    \Delta^{M} \coloneqq \left\{ P \in [0,1]^M \colon \sum_{i=1}^M P_i = 1 \right\}
\end{equation} 
denotes the probability simplex in $\mathbb{R}^M$.
It maps the sequence of transformed embeddings $(\tilde{e}^{(i)})_{i=1}^n \coloneqq \mathcal{T}(\seq{e})$ to a vector $P\in \Delta^{M}$, where $P_i$ describes the probability of predicting $i\in T$ as the next token. Since the transformed embedding of the last token, i.e., $\tilde{e}^{(n)}$, contains information about the whole input text, a simple approach is to use a linear mapping composed with a softmax layer and define 
\begin{equation}
P \coloneqq (\operatorname{softmax} \circ \, \mathcal{L}^{(M,d)})(\tilde{e}^{(n)}).
\end{equation}

\paragraph{Sampling $\mathcal{S}$.}
There are multiple \emph{sampling} strategies $\mathcal{S}\colon \Delta^M \to T$ to arrive at the final prediction for the next token $t^{(n+1)}$, see, e.g.,~\cite{holtzman2019curious}; the arguably simplest one, so-called \emph{greedy sampling}, predicts the token with the highest probability, i.e., 
\begin{equation}
    t^{(n+1)} = \mathcal{S}(P)\coloneqq \argmax_{i=1, \dots, M} P_i,
\end{equation}
see Figure~\ref{fig:transformer}.
One can then apply the same operations to the extended sequence $\seq{t} = (t^{(i)})_{i=1}^{n+1}$, i.e.,
\begin{equation}
    t^{(n+2)} \coloneqq \left(\mathcal{S} \circ \mathcal{H} \circ \mathcal{T} \circ \mathcal{P} \circ \pr{\mathcal{E}}\right)(\seq{t})
\end{equation}
to iteratively compute further tokens\footnote{There is usually a stopping criterion based, e.g., on a special token or the entropy of $P$.}. Due to the autoregressive structure, this can efficiently be done by caching the previous (intermediate) results and only considering the computations for the new token. 

\subsection{Training}
\label{sec: training}
During training, we transform text corpora into sequences of tokens, such that, for a given sequence $(t_i)_{i=1}^n$, we already know the next token $t_{n+1}$ based on the underlying text. One can thus compute the deviation $D$ between the predicted probabilities $P$ of the next token and the ground-truth $t_{n+1}$, typically using a \emph{cross-entropy loss}; in practice, this procedure can be parallelized to compute average losses across many predictions. Using \emph{automatic-differentiation}, one then computes the derivative $\nabla_\theta D$ of the average loss $D$ with respect to the learnable parameters $\theta \in \mathbb{R}^p$ of the transformer $\mathcal{T}$, the embedding $\mathcal{E}$, the prediction head $\mathcal{H}$ (and the positional encoding $\mathcal{P}$ if it is trainable). Updating the parameter by subtracting a sufficiently small multiple $\lambda \in(0,\infty)$ of the derivative, i.e., $\theta_{k+1} = \theta_{k} - \lambda \nabla_\theta D$, one can iteratively minimize the loss---a method known as \emph{stochastic gradient descent}. 
This is the essential mechanism by which word occurrence probabilities are estimated by training from raw data. 
With substantial engineering efforts, more elaborate versions of such training schemes can be parallelized on large GPU clusters and scaled to immense amounts of data. To get an idea of the dimensions, the largest \llamatwo~model with $p=70\cdot 10^9$ parameters was trained for more than $1.7$ million GPU hours on about $2$ trillion tokens of data from publicly available sources~\cite{touvron2023llama}.

\subsection{Training Costs and Emissions}
Training LLMs, as described in the previous section, is a computationally very intensive process and, therefore, costly to carry out in terms of electricity usage (assuming all the hardware would be in place). However, information about training costs and CO$_2$ emissions is not consistently provided in the literature. Notable exceptions include the \textsc{LaMDA} model. The authors~\cite{thoppilan2022lamda} report that a total of $451$MWh was consumed during training, and, as a result, approximately $26$ tons of CO$_2$ were emitted. Using historic US prices\footnote{\url{https://data.bls.gov/timeseries/APU000072610}} of $0.148$ dollars per kWh, this amounts to a cost of $66,748$ dollars. We note that costs may vary by country and by the energy source used to produce energy~\cite{strubell2019energy}. The GLaM model consumes, when trained on the largest dataset, similarly $456$MWh and emits $40.2$ tons of CO$_2$, which places it thus in a similar category to the LaMDA model in terms of cost and emission.

However, more modern LLMs incur significantly more energy consumption and emissions. For instance, the training of \llamatwo~(using $1.7$ million hours on GPUs with a power consumption of about $400$W) emitted more than $291$ tons of Carbon dioxide equivalent (CO$_2$-eq)~\cite{touvron2023llama}. 
LLM vendors (such as OpenAI) typically do not release information about the costs (either in terms of consumed megawatt-hours or (rented) GPU-hours) of training their models, so only vague estimates are possible, which are nonetheless staggering. For example, training the older-generation \gptthree~model~\cite{brown2020language} was estimated, using GPU-hour prices from that time, to run up costs of approximately $4.6$ million dollars~\cite{li2020technical}. 

\section{LLMs for Mathematics}\label{sec:LLMs4Math}

With the foundations of LLMs now well-established, we turn our attention to their application in supporting professional mathematicians. While mathematicians engage in a broad spectrum of mathematical activities, such as performing simulations, modeling, and computation, we focus on the arguably most important task: the capacity of LLMs to generate mathematical proofs. In this sense, our article differs from~\cite{williamson2023deep}, which reviews other general deep-learning approaches to mathematics.

When using an LLM to assist in the task of theorem proving, the simplest way is to directly prompt the model to prove the statement instead of using it for individual steps or other tasks that will be described below. However, many issues have been found with this approach. 
A primary concern is that mathematical arguments hinge on the precision of logic; a single wrong statement very likely invalidates the entire proof.
Assuming that LLMs have a non-negligible, independent probability of error with each predicted word, the \textit{likelihood of producing a correct proof diminishes exponentially with increasing text length.} 
This was also empirically observed in \cite{hendrycks2021measuring}, where an LLM turned out to have a higher accuracy in solving computation tasks if it was asked to skip intermediate steps.

The autoregressive nature of LLMs introduces another critical issue. Once a statement is made, LLMs typically do not revisit or revise their arguments. \emph{This process diverges significantly from the methodologies of most mathematicians.} Rarely does a mathematician draft a complete and detailed proof in a single attempt. Instead, the process often involves crafting a rough sketch, omitting small steps in which we have confidence, iterating, and refining until the proof reaches completion.

Furthermore, LLMs may construct entirely \emph{valid proofs for questions different from those posed}. Such an instance was exemplified in the introduction when the LLM was prompted for a function that is continuous at only one point. Given the prevalence of similar but distinct problems in training datasets, LLMs are likely to respond to the more commonly encountered variations of a question. In this case, a question for a function that is \emph{dis}continuous at only one point.
Similarly, they may prove a theorem under stronger assumptions than stated without making this explicit.

Finally, being based solely on statistical relations of language, LLMs struggle considerably with \emph{arithmetic problems}: These often occur when an LLM has to complete a task, such as carrying out addition or multiplication (in particular, if the involved numbers are large). 
The reason for this is that no numerical solver is built into LLMs. 
Steps towards overcoming this have recently been made by employing a Toolformer approach~\cite{schick2023toolformer}. An instance of this is the WolframAlpha plugin that is available for \gptfour. 

In summary, when LLMs are used to prove theorems, they are susceptible to a range of errors. These errors were examined in \cite{frieder2023mathematical}, to be discussed in the following chapter.
Consequently, a more collaborative approach, incorporating human expertise, is advisable. 
The following strategies appear to be sensible: 

\begin{itemize}
    \item \textbf{Literature/search engine:}
    The LLM can be prompted to explain a definition, find the established name for a vaguely described concept, or find references for a certain statement. 
    In this context, two crucial considerations arise. 
    First, LLMs are known for generating plausible yet fictitious content. 
    This phenomenon is often referred to as \emph{hallucinations}. 
    Therefore, its answer to our queries needs to be verified. 
    Second, the LLM may exacerbate biases in research. This can occur when an LLM overlooks inadequately cited work, effectively burying it while disproportionately recommending over-hyped articles.

    \item \textbf{Brainstorming/Idea Generation:} 
    An LLM can be asked to provide a high-level idea of how to prove a theorem. While this will not produce the full result, it closely resembles the style of a mathematician. 
    There is, however, no guarantee that this idea will be very insightful or lead to something. 
    Being trained on a large corpus of mathematical arguments, an LLM will likely be biased towards recommending the most standard ideas. This may not be helpful for a mathematician who is already an expert in a specific field.
    However, it could be very valuable for a mathematician trying to enter a new area.

    \item \textbf{Proof-checking:}
    An LLM can be asked to find mistakes in a given proof. While there is no guarantee whatsoever that it will find all errors, the ones it finds can often be immediately confirmed as actual mistakes by a mathematician. This is helpful but not reliable. 
    The LLM will likely focus on syntactical correctness over semantic correctness and hence overlook complex errors.

    \item \textbf{Collaborative writing:}
    An LLM can be asked to provide parts or a sketch of a proof and then, after taking feedback from the user, improve parts, repair errors, and add more details. 
    In~\cite{collins2023evaluating}, the interactive performance of three LLMs (\textsc{InstructGPT}, \chatgpt, and \gptfour) on mathematical queries has been measured on a cohort of users. 
    It was attempted to solve mathematical problems by using these models as an assistant. 
    Asking definitions, general mathematical questions (not strictly related to the problem), and proof steps were the three most common use cases.
    It is important to keep in mind that this approach still is susceptible to introducing errors. The study found that self-assessment of individuals---whether they had correctly solved the problem using an LLM---was not always correct. 
\end{itemize}

The approaches above are sensible for the successful employment of modern multi-purpose LLMs. In addition, we anticipate that LLMs specifically designed to prove theorems will be developed in the future. 
One avenue to achieve this is by combining LLM-generated proofs with interactive theorem provers~\cite{de2015lean, barras1997coq}. First steps in this direction have already been taken and appear to be very promising~\cite{frieder2023llms, first2023baldur,yang2019learning, yang2023leandojo}. 

\section{Measuring LLM Performance on Mathematics}

In~\cite{frieder2023mathematical}, an empirical study was carried out to study the mathematical reasoning abilities of three LLMs which were considered to be state-of-the-art in terms of general performance: Two \chatgpt~versions (9-January-2023 and 30-January-2023) and \gptfour. The prompts used to carry out the evaluation correspond to some of the use cases discussed in Section \ref{sec:LLMs4Math}.
\begin{itemize}[leftmargin=*,labelsep=7mm]
\item[\textbf{(PP)}]\textit{Producing proofs:} Exercises from well-known textbooks (\emph{Probability Theory} by R. Durret~\cite{durrett2019probability}, \emph{Topology} by  J. R. Munkres~\cite{munkres2000topology}, and \emph{Functional Analysis} by W. Rudin~\cite{rudin1991functional}) were fed to the LLMs.
    \item[\textbf{(FH)}] \textit{Filling holes:} Proofs with a gap were given to the LLMs, and they were asked to fill the gap. 
    \item[\textbf{(SE)}] \textit{Acting as a mathematical search engine:} The LLMs were asked to give a definition of concepts such as: \enquote{\emph{What is a Banach space?}}. In addition, they were asked to provide the name of definitions, as in \enquote{\emph{How is a complete normed vector space called?}}. Finally, they were prompted to provide proof ideas used in famous theorems. 
    \item[\textbf{(CO)}] \textit{Computation:} The LLMs were given mathematical tasks in which quantities had to be computed.
\end{itemize}

\begin{figure}[!ht]
    \centering
    \includegraphics[width=0.99\textwidth]{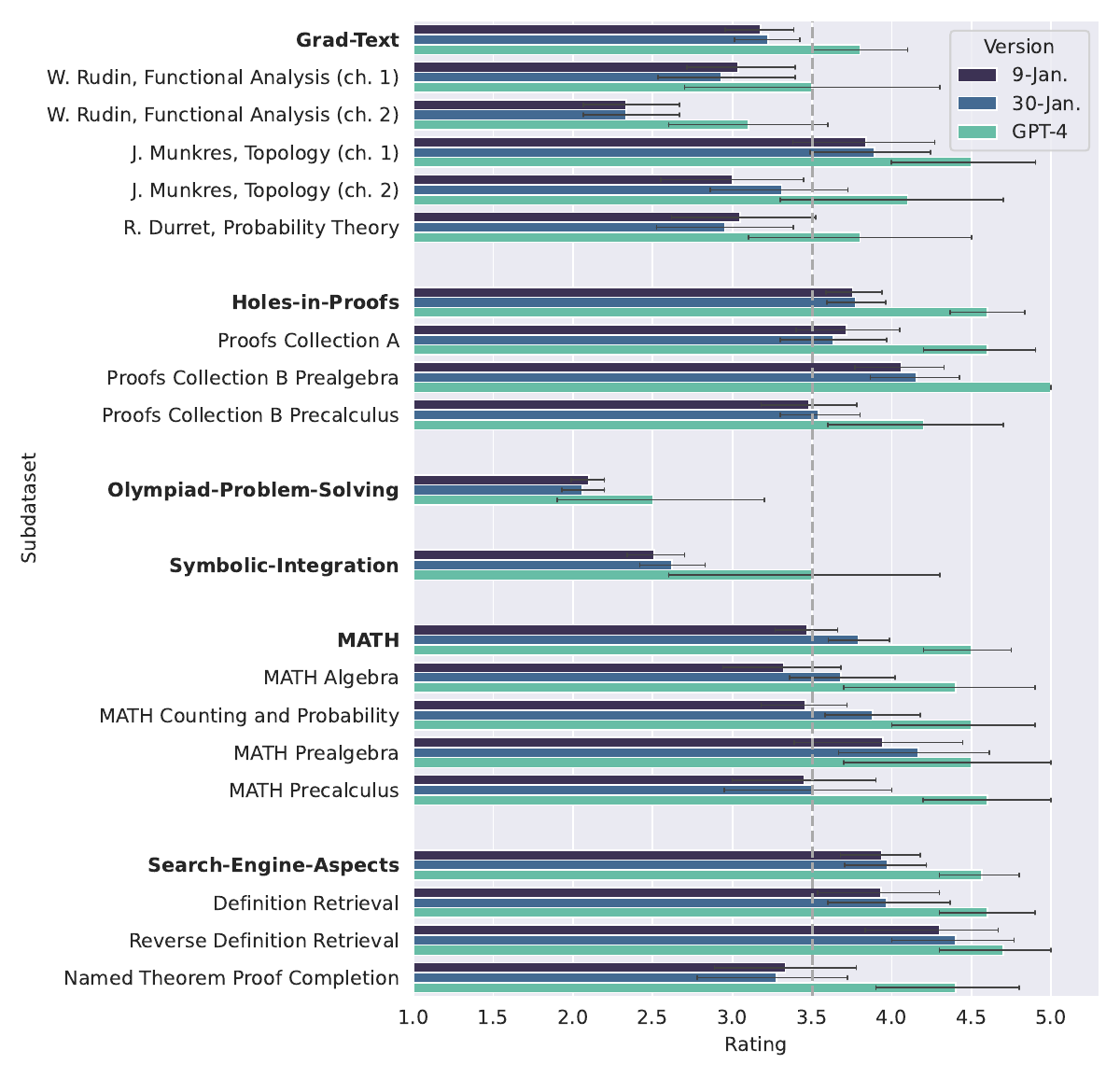}
    \caption{Average rating for each file in each subdataset (bold) of GHOSTS
      for the studied versions of \chatgpt~and \gptfour. The maximal ranking is \texttt{5}, and the minimal ranking, where the question was at least understood, is \texttt{2}; 
       the rating of \texttt{1} indicates that the answer completely misses the question.
       Thus, a reasonable passing grade, i.e., $50\%$ of points, corresponds to a score of \texttt{3.5}, indicated by the dotted line. The error bars represent $95\%$ confidence intervals.}
    \label{fig: AllData}
\end{figure}

To carry out this analysis, the GHOSTS dataset was introduced; see Table~\ref{tab: alldatasets-N} for a more detailed description of its subdatasets (which make up the acronym \enquote{GHOSTS}), as well as how they correspond to the use cases listed above. It consists of 709 prompts, each of which was given to the three considered models. 
The responses of the LLMs were rated by professional mathematicians\footnote{The authors of the present paper form a subset of the evaluators.} on a scale between one (failure to understand the query) and five (perfect or almost perfect output). 
The obtained ratings are shown in Figure \ref{fig: AllData}. We can make the following observations: 

\begin{table}[ht]
\centering
\caption{A summary of all the files from all the subdatasets comprising the GHOSTS dataset, together with their size, i.e., the number of prompts and their associated attribute tags.}
\vspace{1.5em}
\begin{tabular}{lcc} 
\textbf{Subdataset Name} & \textbf{Size} & \textbf{Type}\tabularnewline
\toprule 
\textbf{\emph{G}}\emph{rad-Text} & 130 & PP \tabularnewline
\textbf{\emph{H}}\emph{oles}\emph{-in-Proofs}  & 162 & FH \tabularnewline
\textbf{\emph{O}}\emph{lympiad-Problem-Solving} & 101 & PP \tabularnewline 
\textbf{\emph{S}}\emph{ymbolic-Integration} & 100 & CO \tabularnewline
\emph{MA}\textbf{\emph{T}}\emph{H} & 138 & CO   \tabularnewline
\textbf{\emph{S}}\emph{earch-Engine-Aspects} & 78 &  SE \tabularnewline
\bottomrule
\end{tabular}
\vspace*{2.5em}
\label{tab: alldatasets-N}
\end{table}

\begin{itemize}
    \item \textit{The LLMs work well as a search engine:} \chatgpt~and \gptfour~achieved an excellent score when we asked for definitions of a concept or the name of a theorem or definition. 
    \item \textit{\chatgpt~and \gptfour~struggle with hard questions:} No version of the tested LLMs achieved satisfactory results on the hardest problem set---\emph{Olympiad-Problem-Solving}. 
    Similarly, on the functional analysis questions from Rudin (Chapter 2)---arguably the second most advanced set of questions in the dataset---the results were underwhelming. The ratings were substantially better for more straightforward questions, such as the exercises in topology, which only ask for simple set theory and Boolean logic.
    \item \textit{Good results for simple computations:} Despite not having a built-in numerical solver, \gptfour~performed reasonably well on questions requiring simple computation. For more sophisticated computation involved in \emph{Symbolic-Integration}, \chatgpt~failed and \gptfour~barely achieved a passing grade. 
    \item \textit{User input can have a positive effect:} On the \emph{Holes-in-Proofs} subdataset, we see excellent results in some of the problems. It appears that the additional context given by the user helps the LLMs to produce more truthful solutions. 
    A similar observation was made in a separate experiment where carefully crafted prompts (so-called \emph{prompt engineering}) slightly increased the score of the LLM on the \emph{Olympiad-Problem-Solving} subdataset~\cite{frieder2023mathematical}.
\end{itemize}

\begin{figure}[t]
    \centering
    \includegraphics[width=\textwidth]{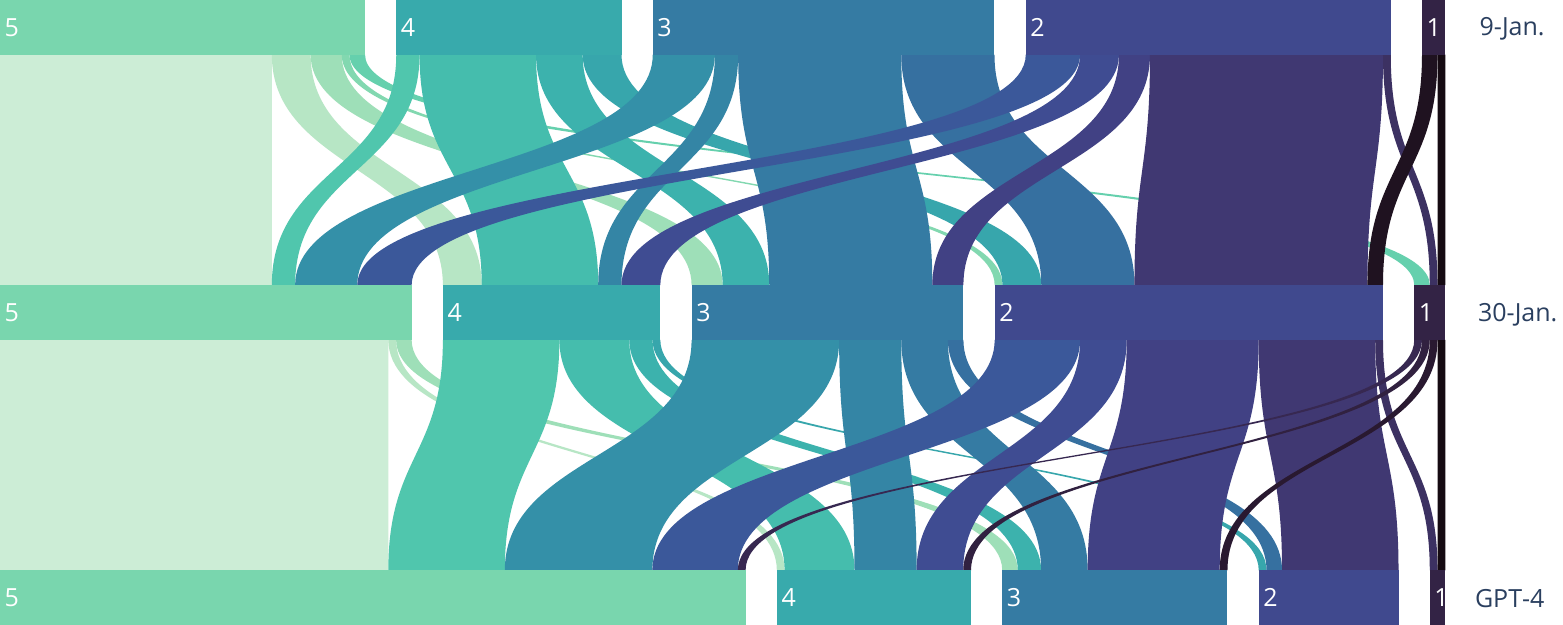}
    \caption{A Sankey diagram of how the ratings evolve from the 9-January-2023 version of \chatgpt~to the 30-January-2023 version \chatgpt{}, and subsequently to \gptfour~(from top to bottom), with all models evaluated on a representative subset of GHOSTS. Each score is color-coded, and the line widths are proportional to the number of ratings.
   }
    \label{fig: sankey}
\end{figure}

The \emph{improvement in rating, as the models become more sophisticated}, is indicated in the Sankey diagram in Figure~\ref{fig: sankey}, which shows how ratings change from one model version to another. We use a representative subset of GHOSTS to advise the Sankey diagram. We observe that between the 9-January-2023 version and the 30-January-2023 version, scores are approximately shuffled, and no substantial increase of the net score occurs. For the newer generation, i.e., \gptfour{}, we can observe a significant improvement in the ratings. This supports the general trend that more modern models also perform better on challenging mathematical reasoning tasks.

In~\cite{frieder2023mathematical}, a subdivision of each question type (as indicated in~\Cref{fig: AllData}) was made, and a much more fine-grained benchmark was introduced that also differentiates potential failure modes. We refer the reader to~\cite{frieder2023mathematical} for further information on more detailed analyses. 

\section{Conclusion}

Our study has highlighted how LLMs possess a remarkable capacity to assist mathematicians in various ways, from detecting and filling gaps in theorems to acting as search engines and finding definitions from descriptions of mathematical objects. 
We have shed light on the inner mechanism of the core piece of architecture that powers modern LLMs, the transformer, and how the way they produce an answer to mathematical questions differs starkly from human reasoning. 

The ability of LLMs to interact with users in a natural language format has made mathematics more accessible, allowing for a broader range of individuals to engage with mathematical research and education. While the full potential of LLMs in automating mathematics is yet to be realized, our findings suggest a promising synergy between human mathematicians and artificial intelligence. High exposure of the work of mathematicians to the effects of LLMs has also been reported in~\cite{eloundou2023gpts}. However, we want to caution that, currently, LLMs are not on a trajectory to replace mathematicians. 
In~\cite{frieder2023mathematical}, it was shown that even the best-performing model has trouble with mathematics on upper-undergraduate difficulties, such as when it is tasked to solve exercises from W. Rudin's \emph{Functional Analysis}~\cite{rudin1991functional}. 
The performance of LLMs has been reported to be below that of humans also in related domains, such as coding challenges in computer science~\cite{koubaa2023humans}. Nonetheless, we anticipate that the emergence of LLMs will be a challenge for education and research. 
Simple exercises or homework and individual steps in mathematical research will be gradually supported by automation or become obsolete. 

\section*{Acknowledgements}
S. Frieder and T. Lukasiewicz were partially supported by the AXA Research Fund.
\bibliographystyle{abbrv}
\bibliography{refs}

\begin{thebibliography}{10}

\bibitem{ba2016layer}
J.~L. Ba, J.~R. Kiros, and G.~E. Hinton.
\newblock Layer normalization.
\newblock {\em arXiv preprint arXiv:1607.06450}, 2016.

\bibitem{barras1997coq}
B.~Barras, S.~Boutin, C.~Cornes, J.~Courant, J.-C. Filliatre, E.~Gimenez, H.~Herbelin, G.~Huet, C.~Munoz, C.~Murthy, et~al.
\newblock {\em The Coq proof assistant reference manual: Version 6.1}.
\newblock PhD thesis, Inria, 1997.

\bibitem{bengio2000neural}
Y.~Bengio, R.~Ducharme, and P.~Vincent.
\newblock A neural probabilistic language model.
\newblock {\em Advances in Neural Information Processing Systems}, 13, 2000.

\bibitem{brown2020language}
T.~Brown, B.~Mann, N.~Ryder, M.~Subbiah, J.~D. Kaplan, P.~Dhariwal, A.~Neelakantan, P.~Shyam, G.~Sastry, A.~Askell, et~al.
\newblock Language models are few-shot learners.
\newblock {\em Advances in neural information processing systems}, 33:1877--1901, 2020.

\bibitem{collins2023evaluating}
K.~M. Collins, A.~Q. Jiang, S.~Frieder, L.~Wong, M.~Zilka, U.~Bhatt, T.~Lukasiewicz, Y.~Wu, J.~B. Tenenbaum, W.~Hart, et~al.
\newblock Evaluating language models for mathematics through interactions.
\newblock {\em arXiv preprint arXiv:2306.01694}, 2023.

\bibitem{de2015lean}
L.~de~Moura, S.~Kong, J.~Avigad, F.~Van~Doorn, and J.~von Raumer.
\newblock The {Lean} theorem prover (system description).
\newblock In {\em Automated Deduction-CADE-25: 25th International Conference on Automated Deduction}, pages 378--388, 2015.

\bibitem{devlin2018bert}
J.~Devlin, M.-W. Chang, K.~Lee, and K.~Toutanova.
\newblock {BERT}: Pre-training of deep bidirectional transformers for language understanding.
\newblock {\em arXiv preprint arXiv:1810.04805}, 2018.

\bibitem{du2022glam}
N.~Du, Y.~Huang, A.~M. Dai, S.~Tong, D.~Lepikhin, Y.~Xu, M.~Krikun, Y.~Zhou, A.~W. Yu, O.~Firat, et~al.
\newblock {GLaM}: Efficient scaling of language models with mixture-of-experts.
\newblock In {\em International Conference on Machine Learning}, pages 5547--5569. PMLR, 2022.

\bibitem{durrett2019probability}
R.~Durrett.
\newblock {\em Probability: Theory and Examples}.
\newblock Cambridge Series in Statistical and Probabilistic Mathematics. Cambridge University Press, 2019.

\bibitem{eloundou2023gpts}
T.~Eloundou, S.~Manning, P.~Mishkin, and D.~Rock.
\newblock {GPTs} are {GPTs}: An early look at the labor market impact potential of large language models.
\newblock {\em arXiv preprint arXiv:2303.10130}, 2023.

\bibitem{first2023baldur}
E.~First, M.~N. Rabe, T.~Ringer, and Y.~Brun.
\newblock Baldur: whole-proof generation and repair with large language models.
\newblock {\em arXiv preprint arXiv:2303.04910}, 2023.

\bibitem{frieder2023llms}
S.~Frieder, M.~Alawadhi, Trimmel, Rashid, and K.~Gy.
\newblock {LLM} vs {ITP}.
\newblock In {\em The 3rd Workshop on Mathematical Reasoning and AI at NeurIPS'23}, 2023.

\bibitem{frieder2023mathematical}
S.~Frieder, L.~Pinchetti, R.-R. Griffiths, T.~Salvatori, T.~Lukasiewicz, P.~C. Petersen, A.~Chevalier, and J.~Berner.
\newblock Mathematical capabilities of {ChatGPT}.
\newblock In {\em Advances in Neural Information Processing Systems}, volume~36, 2023.

\bibitem{gao2020pile}
L.~Gao, S.~Biderman, S.~Black, L.~Golding, T.~Hoppe, C.~Foster, J.~Phang, H.~He, A.~Thite, N.~Nabeshima, et~al.
\newblock The {P}ile: An 800{GB} dataset of diverse text for language modeling.
\newblock {\em arXiv preprint arXiv:2101.00027}, 2020.

\bibitem{haviv2022transformer}
A.~Haviv, O.~Ram, O.~Press, P.~Izsak, and O.~Levy.
\newblock Transformer language models without positional encodings still learn positional information.
\newblock {\em arXiv preprint arXiv:2203.16634}, 2022.

\bibitem{hendrycks2021measuring}
D.~Hendrycks, C.~Burns, S.~Kadavath, A.~Arora, S.~Basart, E.~Tang, D.~Song, and J.~Steinhardt.
\newblock Measuring mathematical problem solving with the {MATH} dataset.
\newblock {\em arXiv preprint arXiv:2103.03874}, 2021.

\bibitem{hendrycks2016gaussian}
D.~Hendrycks and K.~Gimpel.
\newblock Gaussian error linear units ({GELUs}).
\newblock {\em arXiv preprint arXiv:1606.08415}, 2016.

\bibitem{holtzman2019curious}
A.~Holtzman, J.~Buys, L.~Du, M.~Forbes, and Y.~Choi.
\newblock The curious case of neural text degeneration.
\newblock {\em arXiv preprint arXiv:1904.09751}, 2019.

\bibitem{koubaa2023humans}
A.~Koubaa, B.~Qureshi, A.~Ammar, Z.~Khan, W.~Boulila, and L.~Ghouti.
\newblock Humans are still better than {ChatGPT}: Case of the { IEEEXtreme} competition.
\newblock {\em arXiv preprint arXiv:2305.06934}, 2023.

\bibitem{li2020technical}
C.~Li.
\newblock {OpenAI}'s {GPT}-3 language model: A technical overview, 2020.
\newblock \url{https://lambdalabs.com/blog/demystifying-gpt-3}.

\bibitem{liu2021multi}
L.~Liu, J.~Liu, and J.~Han.
\newblock Multi-head or single-head? an empirical comparison for transformer training.
\newblock {\em arXiv preprint arXiv:2106.09650}, 2021.

\bibitem{liu2019roberta}
Y.~Liu, M.~Ott, N.~Goyal, J.~Du, M.~Joshi, D.~Chen, O.~Levy, M.~Lewis, L.~Zettlemoyer, and V.~Stoyanov.
\newblock {RoBERTa}: A robustly optimized {BERT} pretraining approach.
\newblock {\em arXiv preprint arXiv:1907.11692}, 2019.

\bibitem{min2023recent}
B.~Min, H.~Ross, E.~Sulem, A.~P.~B. Veyseh, T.~H. Nguyen, O.~Sainz, E.~Agirre, I.~Heintz, and D.~Roth.
\newblock Recent advances in natural language processing via large pre-trained language models: A survey.
\newblock {\em ACM Computing Surveys}, 56(2):1--40, 2023.

\bibitem{munkres2000topology}
J.~R. Munkres.
\newblock {\em Topology}.
\newblock Prentice-Hall, 2000.

\bibitem{openai2022chatgpt}
OpenAI.
\newblock Introducing {ChatGPT}, 2022.
\newblock \url{https://openai.com/blog/chatgpt}.

\bibitem{openai2023gpt}
OpenAI.
\newblock {GPT}-4 technical report.
\newblock {\em arXiv preprint 2303.0877}, 2023.

\bibitem{ouyang2022training}
L.~Ouyang, J.~Wu, X.~Jiang, D.~Almeida, C.~Wainwright, P.~Mishkin, C.~Zhang, S.~Agarwal, K.~Slama, A.~Ray, et~al.
\newblock Training language models to follow instructions with human feedback.
\newblock {\em Advances in Neural Information Processing Systems}, 35:27730--27744, 2022.

\bibitem{radford2018improving}
A.~Radford, K.~Narasimhan, T.~Salimans, I.~Sutskever, et~al.
\newblock Improving language understanding by generative pre-training, 2018.
\newblock \url{https://openai.com/research/language-unsupervised}.

\bibitem{radford2019language}
A.~Radford, J.~Wu, R.~Child, D.~Luan, D.~Amodei, and I.~Sutskever.
\newblock Language models are unsupervised multitask learners, 2019.
\newblock \url{https://github.com/openai/gpt-2}.

\bibitem{ren2023pangu}
X.~Ren, P.~Zhou, X.~Meng, X.~Huang, Y.~Wang, W.~Wang, P.~Li, X.~Zhang, A.~Podolskiy, G.~Arshinov, et~al.
\newblock {PanGu}-{$\Sigma$}: Towards trillion parameter language model with sparse heterogeneous computing.
\newblock {\em arXiv preprint arXiv:2303.10845}, 2023.

\bibitem{rudin1991functional}
W.~Rudin.
\newblock {\em Functional analysis}.
\newblock McgGraw-Hill, 1991.

\bibitem{sanh2019distilbert}
V.~Sanh, L.~Debut, J.~Chaumond, and T.~Wolf.
\newblock {DistilBERT}, a distilled version of {BERT}: smaller, faster, cheaper and lighter.
\newblock {\em arXiv preprint arXiv:1910.01108}, 2019.

\bibitem{schick2023toolformer}
T.~Schick, J.~Dwivedi-Yu, R.~Dess{\`\i}, R.~Raileanu, M.~Lomeli, L.~Zettlemoyer, N.~Cancedda, and T.~Scialom.
\newblock Toolformer: Language models can teach themselves to use tools.
\newblock {\em arXiv preprint arXiv:2302.04761}, 2023.

\bibitem{sennrich2015neural}
R.~Sennrich, B.~Haddow, and A.~Birch.
\newblock Neural machine translation of rare words with subword units.
\newblock {\em arXiv preprint arXiv:1508.07909}, 2015.

\bibitem{strubell2019energy}
E.~Strubell, A.~Ganesh, and A.~McCallum.
\newblock Energy and policy considerations for deep learning in {NLP}.
\newblock {\em arXiv preprint arXiv:1906.02243}, 2019.

\bibitem{tenney2019bert}
I.~Tenney, D.~Das, and E.~Pavlick.
\newblock {BERT} rediscovers the classical {NLP} pipeline.
\newblock {\em arXiv preprint arXiv:1905.05950}, 2019.

\bibitem{thoppilan2022lamda}
R.~Thoppilan, D.~De~Freitas, J.~Hall, N.~Shazeer, A.~Kulshreshtha, H.-T. Cheng, A.~Jin, T.~Bos, L.~Baker, Y.~Du, et~al.
\newblock La{MDA}: Language models for dialog applications.
\newblock {\em arXiv preprint arXiv:2201.08239}, 2022.

\bibitem{touvron2023llama1}
H.~Touvron, T.~Lavril, G.~Izacard, X.~Martinet, M.-A. Lachaux, T.~Lacroix, B.~Rozi{\`e}re, N.~Goyal, E.~Hambro, F.~Azhar, et~al.
\newblock Llama: Open and efficient foundation language models.
\newblock {\em arXiv preprint arXiv:2302.13971}, 2023.

\bibitem{touvron2023llama}
H.~Touvron, L.~Martin, K.~Stone, P.~Albert, A.~Almahairi, Y.~Babaei, N.~Bashlykov, S.~Batra, P.~Bhargava, S.~Bhosale, et~al.
\newblock Llama 2: Open foundation and fine-tuned chat models.
\newblock {\em arXiv preprint arXiv:2307.09288}, 2023.

\bibitem{vaswani2017attention}
A.~Vaswani, N.~Shazeer, N.~Parmar, J.~Uszkoreit, L.~Jones, A.~N. Gomez, {\L}.~Kaiser, and I.~Polosukhin.
\newblock Attention is all you need.
\newblock {\em Advances in Neural Information Processing Systems}, 30, 2017.

\bibitem{williamson2023deep}
G.~Williamson.
\newblock Is deep learning a useful tool for the pure mathematician?
\newblock {\em arXiv preprint arXiv:2304.12602}, 2023.

\bibitem{wolf-etal-2020-transformers}
T.~Wolf, L.~Debut, V.~Sanh, J.~Chaumond, C.~Delangue, A.~Moi, P.~Cistac, T.~Rault, R.~Louf, M.~Funtowicz, J.~Davison, S.~Shleifer, P.~von Platen, C.~Ma, Y.~Jernite, J.~Plu, C.~Xu, T.~L. Scao, S.~Gugger, M.~Drame, Q.~Lhoest, and A.~M. Rush.
\newblock Transformers: State-of-the-art natural language processing.
\newblock In {\em Proceedings of the 2020 Conference on Empirical Methods in Natural Language Processing: System Demonstrations}, 2020.

\bibitem{yang2019learning}
K.~Yang and J.~Deng.
\newblock Learning to prove theorems via interacting with proof assistants.
\newblock In {\em International Conference on Machine Learning}, pages 6984--6994. PMLR, 2019.

\bibitem{yang2023leandojo}
K.~Yang, A.~M. Swope, A.~Gu, R.~Chalamala, P.~Song, S.~Yu, S.~Godil, R.~Prenger, and A.~Anandkumar.
\newblock {LeanDojo}: Theorem proving with retrieval-augmented language models.
\newblock {\em arXiv preprint arXiv:2306.15626}, 2023.

\bibitem{zhao2023survey}
W.~X. Zhao, K.~Zhou, J.~Li, T.~Tang, X.~Wang, Y.~Hou, Y.~Min, B.~Zhang, J.~Zhang, Z.~Dong, et~al.
\newblock A survey of large language models.
\newblock {\em arXiv preprint arXiv:2303.18223}, 2023.

\end{thebibliography}

\end{document}